\def\year{2022}\relax
\documentclass[letterpaper]{article} 
\usepackage{aaai21}  
\usepackage{times}  
\usepackage{helvet} 
\usepackage{courier}  
\usepackage[hyphens]{url}  
\usepackage{graphicx} 
\usepackage{natbib}  
\usepackage{caption} 
\usepackage{enumitem}
\usepackage{color}
\usepackage{textcomp}
\usepackage{subfigure}
\usepackage{xspace}
\usepackage{booktabs}
\usepackage{colortbl}
\usepackage{setspace}
\usepackage{csquotes}
\usepackage{longtable}
\usepackage{microtype}
\usepackage{placeins}
\usepackage{xcolor}
\usepackage{bbding}
\usepackage{epsfig}
\usepackage{dirtytalk}
\usepackage{multirow}
\usepackage{tabularx}
\usepackage{array}
\usepackage{soul}
\usepackage{amssymb}
\usepackage[normalem]{ulem}
\usepackage{siunitx}
\usepackage{amsmath}

\definecolor{Orange}{rgb}{1,0.5,0}
\definecolor{Purple}{rgb}{1,0,1}
\definecolor{Red}{rgb}{1,0.3,0.3}
\definecolor{Blue}{rgb}{0,0,1}
\definecolor{Green}{rgb}{0,0.68,0}

\newcommand{\todo}[1]{\textsf{\textbf{\small \textcolor{Orange}{[[#1]]}}}}
\newcommand{\mm}[1]{\todo{$\stackrel{\mbox{\tiny MM}}{\mbox{\tiny says}}$:\;#1}}

\newcommand{\new}[1]{#1}

\newcommand{\review}[1]{#1} 

\title{``Dummy Grandpa, do you know anything?":\\ Identifying and Characterizing Ad hominem Fallacy Usage in the Wild}

\newcommand{\textincon}[1]{{\fontfamily{zi4}\selectfont #1}} 

\begin{document}
\date{}

%

\author{%
Utkarsh Patel, Animesh Mukherjee, Mainack Mondal\\
\normalsize \normalfont{Indian Institute of Technology, Kharagpur}\\
\normalsize \normalfont{utkarshpatel@iitkgp.ac.in, animeshm@cse.iitkgp.ac.in, mainack@cse.iitkgp.ac.in}%
}

    


\maketitle


\begin{abstract}
Today, participating in discussions on online forums is extremely commonplace and these discussions have started rendering a strong influence on the overall opinion of online users. Naturally, twisting the flow of the argument can have a strong impact on the minds of na\"ive users\new{,} which in the long run might have socio-political ramifications, for example, winning an election or spreading targeted misinformation. Thus, these platforms are potentially highly vulnerable to malicious players who might act individually or as a cohort to breed fallacious arguments with a motive to sway public opinion. \textit{Ad hominem} arguments are one of the most effective forms of such fallacies. Although a simple fallacy, it is effective enough to sway public debates in offline world and can be used as a precursor to shutting \new{down} the voice of opposition by slander. 

In this work, we take a first step in shedding light on the usage of ad hominem fallacies in the wild. First, we build a powerful ad hominem detector based on transformer architecture with high accuracy (F1 more than $83\%$, showing a significant improvement over prior work), even for datasets for which annotated instances constitute a very small fraction. 
We then used our detector on $265$k arguments collected from the online debate forum – \textit{CreateDebate}. Our crowdsourced surveys validate our in-the-wild predictions on CreateDebate data ($94\%$ match with manual annotation). Our analysis revealed that a surprising $31.23\%$ of CreateDebate content contains ad hominem fallacy, and a cohort of highly active users post significantly more ad hominem to suppress opposing views. Then, our temporal analysis revealed that ad hominem argument usage increased significantly since the $2016$ US Presidential election, not only for topics like Politics, but also for Science and Law. 
We conclude by discussing important implications of our work to detect and defend against ad hominem fallacies. 

\end{abstract}

\section{Introduction}

Today online forums and social media sites facilitate easy collaborative opinion formation for billions of users surpassing geographical boundaries. However, perhaps quite naturally, this process of opinion formation also involves participating in online arguments where multiple parties often present their conflicting views. The caveat here is that the arguments presented in online debates are not always sound. They often contain \textit{deceptive arguments in disguise}~\citep{Kennedy1993-KENAOR-4}. Intuitively, in online forums, the users present informal fallacies, necessitating \new{deep} content analysis to identify them (as opposed to the formal ones, which can be examined using logical representations)~\citep{sahai-etal-2021-breaking}. 

Amongst different fallacies, \textit{ad hominem} is perhaps the most famous one in the offline world~\citep{Macagno2013StrategiesOC,Schiappa2013,Zalta2004-ZALTSE, woodsetal}. Ad hominem or \textit{against the person} is a fallacious argument, based on feelings of bias (mostly irrelevant to the argumentation), rather than reality, reason, and rationale. However, despite a long history of dissecting and condemning ad hominem fallacies in the offline world, even online users are no stranger to the usage of ad hominem fallacies~\citep{adhominem2,adhominem5}. Ad hominem arguments are often personal attacks on someone’s character or motive rather than an attempt to address the reasoning that they presented. People tend to use ad hominem arguments because they want to appeal to others’ emotions rather than reasoning.

Recently, there has been substantial research concerning investigating and countering hate speech, misinformation as well as cyberbullying within the user-generated content posted on social media~\citep{mondal-2017-hatespeech,mondal-2018-hatespeech-journal,das2021brutus,mathew2020hate,mathew2020hatexplain}. In the same vein, although relatively \new{scarce}, some very recent works are exploring the detection of ad hominem fallacies in the wild using computational methods~\citep{habernal-etal-2018-name,sahai-etal-2021-breaking}. However, these works focused more on the detection of ad hominem (and other) fallacies using automated methods in online forums. There is not much work shedding light on the lay of the land for ad hominem usage over time. We aim to bridge this gap. \new{Specifically, we ask the following research questions:}

\begin{enumerate}
	
	\item \new{Can we design a practical ad-hominem detector which can uncover ad-hominems in the wild with high accuracy?}
	
	\item \new{How does the dynamics of ad hominem argumentation evolve with time in the wild? Who are users that play the key role in posting these ad hominem arguments?}

\end{enumerate}

\new{To address these questions, in} this work, we present a data-driven exploration of ad hominem arguments in the wild using CreateDebate\footnote{\url{https://www.createdebate.com/}}, an online discussion forum, as an experimental testbed. We used an in-house high-accuracy and high external validity ad hominem detector on a dataset containing more than $18$k posts with $265$k comments generated by $15$k users of CreateDebate. Next, we analyzed the detected large-scale ad hominem arguments to shed light on in-the-wild ad hominem usage. Specifically, we have made three key contributions to this work.

First, \new{to answer the first research question}, we developed ad hominem detectors considering two scenarios---when the annotated data is abundant and when it is not, using an annotated dataset from previous work on Reddit~\cite{habernal-etal-2018-name}. Our models significantly improved over the ad hominem detectors reported in prior work and achieved a macro-F1 score more than $0.83$. Furthermore, we evaluated the predictions of our detector on the CreateDebate dataset using a user study. Our user study demonstrated that results from our detector (trained on Reddit data) are also externally valid---achieving a $94\%$ accuracy on the CreateDebate data. \new{Overall, unlike previous work~\cite{habernal-etal-2018-name}, leveraging modern transformer models enabled us to build a high-accuracy detector and effectively handle situations with low amount of annotated data.}

Subsequently, \new{to answer the second research question}, we leveraged this detector on our large-scale CreateDebate data and found that a significant $31.23\%$ (i.e., almost one-third) of all the comments on CreateDebate are ad hominem. On further investigation, we uncovered that a community of highly active users posts a disproportionately high volume of ad hominem fallacies (more than $50\%$ of their total comments). 

Third, \new{to dig deeper into our second research question}, we checked whether ad hominem arguments were always used in such a high volume, or was it just a recent trend? It appeared that the fraction of ad hominem arguments showed a sharp rise after $2016$. This trend was prominent not only in the Politics subforum, but transcended in subforums like Science and Law. We found a striking correlation--—the userbase of the Politics subforum of CreateDebate has a high overlap with userbase for subforums like Science and Law. The rise of ad hominem arguments in Politics subforums seems to be triggered by the $2016$ US Presidential election, which resulted in the users active in Political subforum posting insulting comments in other forums as well, hence, significantly increasing the ad hominem usage. We
also release our model to the research community.\footnote{\url{https://bit.ly/3TJbMPr}}

\vspace{2mm}
\noindent \textbf{Ethical considerations:} In this work, we collected and analyzed data from CreateDebate and also conducted an annotation survey for validating our classifier. However, since we were analyzing user-generated data in this work, we tried our best to conduct our study ethically and protect the privacy of the users in our dataset. Specifically, we leveraged the previous work by \citet{eysenbach_ethical_2001} to check the ethics of our work. We noted that CreateDebate is a moderate-sized forum with around $15$k members, and no registration was necessary to view and collect the CreateDebate data, signifying it was an ‘open’ forum. Finally, the debate topics often revolved around general phenomena (e.g., election), signifying the potentially public nature of our collected dataset. Nonetheless, following the footsteps of previous work by ~\citet{cook-2018-ethics}, we hashed usernames after data collection to protect the privacy of the users during our analysis. Along the same lines of ethical consideration, for our annotation study, we did not collect any personally identifiable data from our participants to protect their privacy. In the next section, we shall start with related research to put our work in context.

\section{Related works}

We divide prior works into two important sub-parts---exploration on ad hominem argumentation, especially in online forums like ChangeMyView\footnote{\url{https://www.reddit.com/r/changemyview/}} and CreateDebate, and usage of Generative Adversarial Networks (GANs) in NLP. 

\subsection{Investigation on ad hominem argumentation}

\noindent Aristotle first identified that some arguments are indeed \textit{deceptions in disguise}~\citep{Kennedy1993-KENAOR-4}. He called them fallacies. The ad hominem arguments are addressed in most of the follow-up treatises of fallacies~\citep{Hamblin1970-HAMF-6, VanEemeren1987-VANFIP-2, Boudry2015-BOUTFT}. Ad hominem arguments are used in a debate for simply attacking the opponents’ traits instead of countering their arguments~\citep{Tindale2007-TINFAA-2}. Naturally, ad hominem arguments are based on feelings of bias rather than reason, often involving personal attacks on someone’s character or motive. Though arguing \textit{against the person} is considered faulty, these arguments are used in online debate forums and social media sites~\citep{habernal-etal-2018-name,sahai-etal-2021-breaking}. Ad hominem arguments are multifaceted and use complex strategies, involving not a simple argument, but a cumulation of several combined tactics~\citep{Macagno2013StrategiesOC}. \new{Interestingly, the issue in some type of ad hominem arguements is not logical inconsistency, but rather what one would call a pragmatic inconsistency. It refers to a kind of inconsistency between asserted
statements and personal actions \citep{walton1998ad}. However, this type of inconsistency is out of scope for this work.} Specifically, the majority of the above research was often aimed at dissecting what constitutes ad hominem in philosophy, rather than evaluating its usage in the real world~\citep{Schiappa2013, Macagno2013StrategiesOC, Zalta2004-ZALTSE, woodsetal}. 

Recently, the scenario started to change when researchers working on NLP aimed to identify different fallacies in online forums. To that end, \citet{10.1145/3038912.3052591} annotated $38$k instances of Wikipedia talk page comments for detecting personal attacks on the forum and \citet{JAIN14.1019} studied principal roles in discussions from the Wikipedia Article for Deletion pages, and extracted several typical roles like ‘idiots’, ‘voices’, ‘rebels’, etc. which might be considered signals for ad hominem fallacies.

\noindent\textbf{Studies on ChangeMyView}: More recent work exploited online discussion forums like Reddit, primarily the ChangeMyView subreddit, to detect naturally occurring hate-speech and ad hominem fallacies.
\citet{wei-etal-2016-post} studied the impact of different sets of features on the identification of persuasive comments. \citet{tan2016} developed a framework for analyzing persuasive arguments and malleable opinions. \citet{habernal-etal-2018-name} investigated ad hominem argumentation at three levels of discourse complexity (arguments in isolation, in direct replies to original post without dialogical context and in a larger inter-personal discourse context). \citet{sahai-etal-2021-breaking} extended this work and found the types of fallacies. Our work builds on this type of detection methods, yet extends them considerably. 

\noindent\textbf{Studies on CreateDebate}: \citet{abbott-etal-2016-internet} developed Internet Argument Corpus (v$2.0$), a collection of corpora for research in political debate on Internet forums, which contains a sample from CreateDebate ($3$k posts) and includes topic annotations. \citet{wei-etal-2016-preliminary} analyzed the disputation action in the online debate by labeling a set of disputing argument pairs extracted from CreateDebate and performing annotation studies. \citet{trabelsi-zaiane-2014-finding} suggested a fine-grained probabilistic framework for improving the quality of opinion mining from online contention texts. \citet{hasan-ng-2014-taking} exploited stance information for reason classification, proposing systems of varying complexity for modeling stances and reasons. \citet{QIUphdthesis} modeled user posting behaviors and user opinions for viewpoint discovery and proposed an integrated
model that jointly considers arguments, stances, and attributes. \citet{Qiu2015ModelingUA} predicted user stances on a variety of topics and assembled user arguments, interactions, and attributes into a collaborative filtering framework that exploits recently introduced fast inference methods.


\subsection{Text classification using deep learning for our detector}

We explored a number of state-of-the-art transformer architectures like Google’s BERT \citep{devlin2019bert}, OpenAI's GPT-1 \citep{Radford2018ImprovingLU} for our experiments. These architectures are trained over large-scale annotated datasets and only require fine-tuning for a targeted task to achieve high accuracy for various NLP applications. However, one major disadvantage of these architectures is that even fine-tuning them often requires at least thousands of annotated examples for the targeted tasks. However, in our use case, obtaining thousands of annotated ad hominem arguments (and an equal number of non-ad hominem arguments) is costly and might be difficult to obtain. To that end, we leveraged a recent technique—integrating these huge pre-trained models with Generative Adversarial Networks (GANs)~\citep{10.5555/2969033.2969125}. In GANs, a `generator' is trained to produce samples resembling some data distribution. This training process ‘adversarially’ depends on a ‘discriminator’, which instead is trained to distinguish examples of the generator from the real instances. SS-GANs are an extension to GANs where the discriminator also assigns class labels to each example while discriminating whether or not it was naturally produced~\citep{NIPS2016_8a3363ab}. \citet{croce-etal-2020-gan} proposed `GAN-BERT' that extends the fine-tuning of BERT-like architectures with unlabeled data in a generative adversarial setting using SS-GAN schema. Building on this related work helped us create an accurate yet explainable detector with limited data. Next, we will start with describing our approach to develop the classifier. 

\vspace{2mm}

\noindent\textbf{Present work}: Our work builds a highly accurate detector, beating the models used by \citet{habernal-etal-2018-name} in macro averaged F1-score for classification and establishes the external validity of the detector on CreateDebate data. 
\new{We created a novel, significantly improved ad hominem detector and demonstrated its  utility in detecting adhominem in the wild (with very limited annotated sample, which considerably increased the utility of the model).}
Then using this detector, we measured the prevalence of ad hominem in the wild—we found that the amount of ad hominem in recent times increased manyfold, more than the figures hinted in any of the previous works. Our highly accurate detector relied on the recent advances in text classification using transformer models as discussed above. \new{Our analysis revealed \textit{for the first time}, the extremely worrying prevalence of ad hominem in the wild---one-third of the posts in the CreateDebate forum were ad hominems and a small cohort of highly active
users hurl the largest number of ad hominems. Quite interestingly, hurling ad hominems accelerated at time periods closer to the $2016$ US Presidential election. Overall, political debates are found to be at the core of increased ad hominem usage with its effects transcending to other topics like religion, science and law.}

\section{Ad hominem detection}\label{sec:detection}

For our experiments, we used ad hominem argumentation in the ChangeMyView (CMV) dataset \citep{habernal-etal-2018-name} as benchmark. ChangeMyView is a popular subreddit in which a user (called OP, original poster) posts an opinion and other users write comments to change the perspective of OP about the posted opinion. OP can acknowledge convincing arguments by giving \textit{delta} points.

Unlike regular debate forums, strict rules are enforced on this subreddit content. \new{Violating these rules results in deletion of the content by moderators.} The CMV dataset contains $7242$ comments from this subreddit ($3622$ instances with the label ‘ad hominem’ and $3620$ instances with the label ‘none’). The dataset was created maintaining a balance of syntactic and semantic similarity between the instances of the two-class labels.
\new{We refer the reader to \cite{habernal-etal-2018-name} for additional details about the dataset.}
We will use this dataset to build powerful classifiers for ad hominem detection.

\new{\citet{habernal-etal-2018-name} used CNN and $2$ Stacked Bi-LSTM models for detecting ad hominem comments. They reported $10$-fold cross validation results.} We
used
the BERT model (case-insensitive, base)
\new{and carried out the same $10$-fold cross validation experiments. The results are presented in Table~\ref{table:BERTvsBaselines}.}

\begin{table}[htbp]
\centering
\small
\begin{tabular}{lrr}
\hline
\textbf{Model} & \textbf{Accuracy} & \textbf{Macro-F1}  \\ \hline
CNN           & \new{$0.808$}   &\new{$0.807$}                     \\
2 Stacked Bi-LSTM         &\new{$0.781$}    &\new{$0.781$}                   \\
BERT           &\new{\textbf{$0.834$}}          &\new{\textbf{$0.834$}}        \\
\hline
\end{tabular}
\caption{
\new{$10$-fold cross validation results on the CMV dataset for classification of ad hominem comments using CNN (baseline), 2 Stacked BiLSTM (baseline) and BERT}.}
\label{table:BERTvsBaselines}
\end{table}

\begin{figure}[]
\centering
\includegraphics[width=\linewidth]{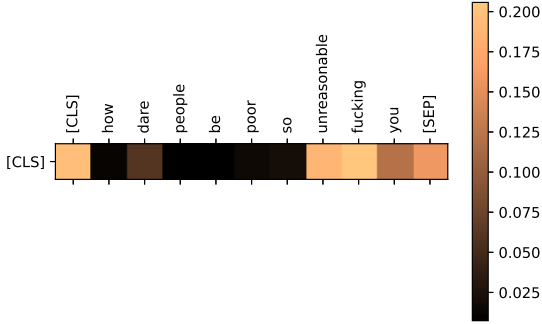}
\caption{Example visualization of attention scores for the \textincon{[CLS]} token. Words like \textincon{unreasonable}, \textincon{fucking}, etc. get more attention, hence, have more influence on the classification.}
\label{fig:AttentionScoreSample}
\end{figure}

\begin{figure*}[htbp]
\centering
\includegraphics[width=\linewidth]{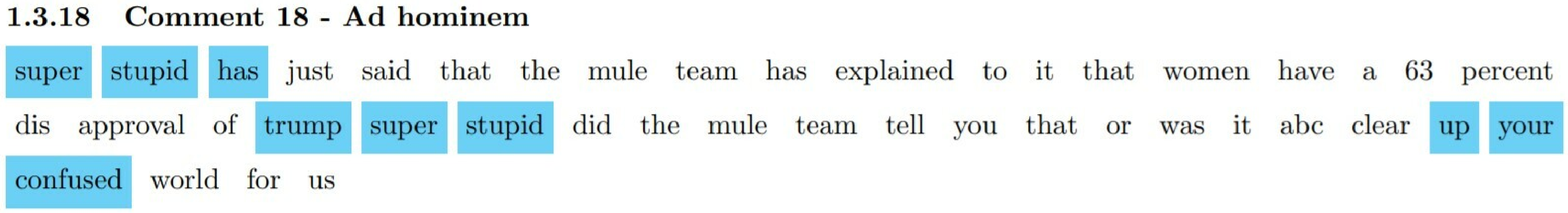} \\
\includegraphics[width=\linewidth]{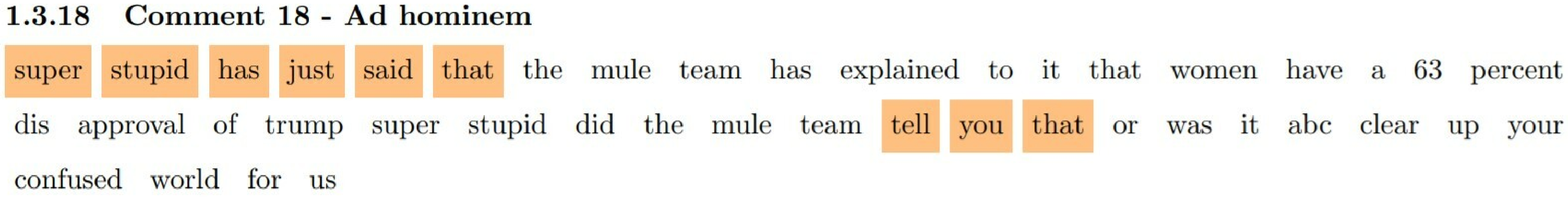} \\
\caption{Example
\new{visualization of trigrams, which have largest influence on classification, based on}
attention scores of \textincon{[CLS]} token for an ad hominem comment using BERT \textit{(cyan)} and GAN-BERT \textit{(orange)}. Even with very less annotated data, GAN-BERT is able to detect trigrams similar to that of BERT.}
\label{fig:WordWeightHeatMapBert}
\end{figure*}

As noted in Table~\ref{table:BERTvsBaselines}, we achieved a \new{$2.6$}\% improvement in
\new{accuracy} over the baselines. We further investigate the segments that could be potentially responsible for flagging an argument as ad hominem.
We
cast an explanation of triggers and dynamics of ad hominem argumentation as a supervised learning problem and draw theoretical insights by a retrospective interpretation of the learned model. \new{As \textincon{[CLS]} token is the aggregate representation of the input sequence for classification tasks} we use attention scores\footnote{\new{In a transformer based architecture, attention scores refer to the overall strength of the relationship of a word with the other words in the sequence. Attention scores over input regions or intermediate features are often interpreted as a contribution of the attended unit to the inference made by the model~\cite{rigo:22}.}} for \textincon{[CLS]} token to identify key tokens influencing the classification. A sample visualization of the attention scores is shown in Figure~\ref{fig:AttentionScoreSample}. We greedily select top three tokens (excluding \textincon{[CLS]} and \textincon{[SEP]} tokens) on the basis of attention scores so that the trigrams centered at those tokens do not overlap and
\new{highlight these trigrams in the visualization}. A sample
\new{visualization of these trigrams} is shown in Figure~\ref{fig:WordWeightHeatMapBert}. We analyzed these
\new{highlighted trigrams} for the comments which were flagged as ad hominem and observed that the BERT model is not only able to beat the baselines in terms of accuracy, but also the highlighted trigams in this case can be easily interpreted to extract linguistic insights into the potential triggers for ad hominem argumentation.

\new{One bottleneck for such studies is the cost of generating an annotated dataset.} Annotating ad hominem arguments is a \new{very} costly task \new{as they are difficult to comprehensively and objectively define \citep{DBLP:journals/corr/abs-2010-12820}}. Hence, we simulate a situation where we assume that labeled instances constitute a very small fraction of the dataset. While doing training on different folds, we intentionally drop the class labels of a given fraction of instances and then fine-tune the BERT model on only the labeled instances and evaluate its performance. This is repeated for different fractions of labeled instances while training. We observe that as we increase the fraction of unlabeled instances, the macro-F1 score for classification degrades. This is a major disadvantage of transformer architectures like BERT---they require thousands of annotated examples to achieve state-of-the-art results on the targeted tasks. However, with \new{few} annotated examples, they fall apart.

\begin{figure}[]
\centering
\includegraphics[width=\linewidth]{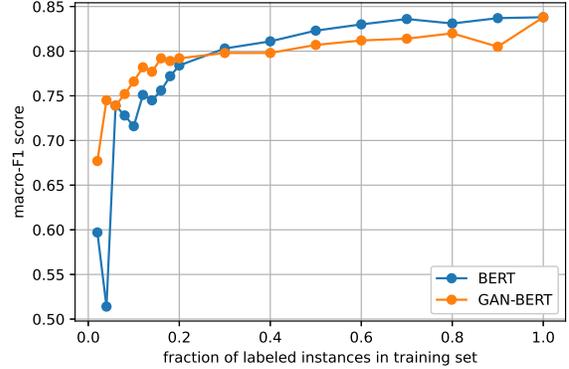}
\caption{BERT vs. GAN-BERT -- Macro-F1 scores for ad hominem classification for different fraction of labeled instances in the training set. When fraction of labeled instances is low, GAN-BERT beats BERT. However, for high fraction, BERT outperforms GAN-BERT.}
\label{fig:BERTvsGANBERT}
\end{figure}

We experimented with GAN-BERT to leverage the unlabeled instances in the training set in a generative adversarial setting, which we simply cannot use while working with BERT. As the results in Figure~\ref{fig:BERTvsGANBERT} show, the requirement of labeled instances can be drastically reduced (up to only $50-100$ labeled instances) for obtaining similar performance as that of BERT. When low fraction of instances in training set are annotated, GAN-BERT beats BERT. One of the possible reasons could be that as the number of annotated instances \new{decreases}, GAN-BERT leverages its GAN architecture to use the unlabeled instances to train its discriminator. As this utility is not present in BERT, its performance is not very good. Naturally, when \new{a} larger fraction of annotated instances are available in the training set, BERT beats GAN-BERT since the utility of the GAN architecture is diminished. 
Figure~\ref{fig:WordWeightHeatMapBert} compares the visualizations obtained by using BERT (when $100$\% training instances are labeled) and GAN-BERT (when $15$\% training instances are labeled). We observe that even with much less annotated data, GAN-BERT is able to detect triggers similar to that of BERT. In the rest of the paper we will report results using the BERT-based model, the GAN-BERT models produced similar results.


\section{Detecting ad hominem fallacies in the wild}\label{sec:createdebatedetection}

\begin{table*}[t]
\centering
\small
\begin{tabular}{p{0.10\linewidth}  p{0.85\linewidth}}
\hline
\textbf{Topic} & \textbf{Examples}    \\ \hline
Politics       & \scriptsize{\textbf{(1) }\textit{STUPOR STUPID you think a lot in the CONFUSED MIND you have ! And by your IGNORANT RESPONSE you further expose you are a consumer of a media that lies to you and you LAP it up like the GOOD LITTLE LEFTIST you are !!!!!}} \\
               & \scriptsize{\textbf{(2) }\textit{Now Socialist have ya not paid any attention to the Gubernor of Michigan ? Damn Socialist are you as stupid as you seem to be when you engage your 1 brain cell}} \\ \hline
Religion       & \scriptsize{\textbf{(1) }\textit{Tell me all about your love for the Muslims you Party Parrot ? Stupid i thought you were a Jew ? Do you have confusion each day until ya turn on DNC Media ?}} \\
               & \scriptsize{\textbf{(2) }\textit{you don't have the right to believe in a sacrifice that never happen, why don't you find an ass to wipe you're a massive sack of shite with half of half a mind you're asinine you think that black is white you're more dishonest then a bag of kykes}} \\ \hline
World News     & \scriptsize{\textbf{(1) }\textit{You threatening to kill anyone who disagrees with Corbyn is fascism, yes. Christ you're such a liar it's embarrassing. Do you actually believe the words you type you boring fucking fascist troll?}} \\
               & \scriptsize{\textbf{(2) }\textit{The Fat Boy Dicktater is launching missiles and is it for you to say the next missile does not carry an ICBM and how would you know ? Are you using a statement to prove a point you no nothing about ?}} \\ \hline
Science        & \scriptsize{\textbf{(1) }\textit{You haven't made a single argument to disprove the existence of a designer .Not even one That is because I am not disputing that there may have been a designer, you retarded imbecile. Do you even read the stuff you reply to?}} \\
               & \scriptsize{\textbf{(2) }\textit{Don't understand even the basics of how chemotherapy works or are anti-science I see. Oh Jesus Christ you are just soooooooooooooooooooooo stupid. UV radiation has got nothing to do with chemotherapy you brainless Nazi retard. Chemotherapy uses gamma radiation to target cancer cells. It also makes patients extremely sick and causes their hair to fall out. You're an idiot and every single word you type is stupid.}} \\ \hline
Law            & \scriptsize{\textbf{(1) }\textit{Ah , the American patriot who never heard of implied consent laws, why not do a bit of reasearch you dummy ? I love it when an idiot like you starts back tracking as you claimed cops could not pull you over now they can. A cop can pull you over for various reasons , you've just done a huge about turn on your lazy assertions as usual}} \\
               & \scriptsize{\textbf{(2) }\textit{Amy anyone can speak in terms you whine about without your sensitive ears hearing what was said. Keep crying u college educated fool because you cannot stop free speech in any forum !}} \\ \hline
Technology     & \scriptsize{\textbf{(1) }\textit{Shut up you bitter old fraud. You're a twisted, delusional, selfish piece of excrement with precisely zero personal integrity. You are a walking, talking form of ass cancer.}} \\\
               & \scriptsize{\textbf{(2) }\textit{Excuse me you imbecile, I graduated from Bismarck State College with a bachelors in Farm and Ranch Management. I was known on campus as ``The Great Debater'', and successfully won 18 arguments. So far I have been gentlemanly, but if you keep up with this funny business you will make me unleash my inner demons and go full throttle debate god.}} \\ \hline
\end{tabular}
\caption{Examples of ad hominem arguments across different topical sub-forums of CreateDebate.}
\label{table:AdHominemExamples}
\end{table*}

Now that we created an accurate and explainable ad hominem detector using the Reddit data, we aimed to test the usage of ad hominem fallacies in discussion forums from the wild. To that end, we chose CreateDebate as our experimental testbed. 

\subsection{Collecting CreateDebate dataset}

CreateDebate is a social networking debate website, launched in $2008$. It was built around ideas, discussion, and democracy to help groups of people to sort through issues, viewpoints, and opinions. The discussions in CreateDebate often aim towards reaching consensus and understanding to make better decisions. CreateDebate is very similar to Reddit with a notable exception---its moderation policy is different, only the debate creator can be the debate moderator\footnote{\label{cdfaq} \url{https://www.createdebate.com/about/faq}}. 

Each CreateDebate post is created by a user who also acts as its \new{sole} moderator. The forum allows users to write their perspectives as comments on the posts. Other users can support a comment, dispute it or clarify it as replies. The site, like Reddit, doesn't limit the depth of comment nesting. CreateDebate forum is divided into $14$ topical forums---Politics, Entertainment, World News, Religion, Law, Science, Technology, Sports, Comedy, Business, Travel, Shopping, Health, and NSFW. The majority of the content in CreateDebate is public for all the forums. We found CreateDebate to be suitable for our investigation since it is a popular discussion-based forum in the wild with \new{weak} moderation. In effect, CreateDebate provides us an opportunity to measure the prominence of ad hominem fallacy usage over time.

\begin{table}[!htbp]
\centering
\small
\begin{tabular}{lrrr}
\hline
\textbf{Topic}      & \textbf{\# Posts} & \textbf{\# Comments} & \textbf{\# Users}   \\ \hline
Politics            & $10,434$            & $119,850$              & $7,686$               \\
Religion            & $2,841$             & $77,418$               & $4,563$               \\
World News          & $2,008$             & $27,418$               & $3,622$               \\
Science             & $1,276$             & $20,691$               & $2,837$               \\
Law                 & $759$               & $11,016$               & $1,436$               \\
Technology          & $909$               & $8,421$                & $2,674$               \\ 
\hline
\textbf{Total}      & $18,227$            & $264,814$              & $14,961$              \\
\hline
\end{tabular}
\caption{Basic statistics of our collected CreateDebate dataset. The first post in our dataset was posted on February $20$, $2008$ and the last post was updated on November $24$, $2021$.}
\label{table:CreateDebateDatasetDescription}
\end{table}


In this work, we programmatically collected the complete publicly available CreateDebate dataset for all topical CreateDebate forums from the inception of the CreateDebate service. However, for brevity, we will present primarily results from the top six CreateDebate forums---Politics, Religion, World News, Science, Law, and Technology. Results from all other topical forums remained the same. We present the general statistics of the dataset in Table~\ref{table:CreateDebateDatasetDescription}---in totality, these six forums contain $18,227$ posts with $264,814$ comments made by $14,961$ users uploaded over $14$ years. We verified that all posts we collected from CreateDebate were in English. We leveraged this large-scale discussion dataset (posted over the years) collected from CreateDebate to detect ad hominem from online discussions. However, we faced a crucial question---is our detector, trained over the Reddit CMV dataset, extendable to the CreateDebate dataset? We explored this question next. 



\subsection{Validating our ad hominem detector on CreateDebate discussions}


After collecting the CreateDebate data, we faced a dilemma---our ad hominem detector was fine-tuned on the Reddit CMV dataset (as noted in the previous section), however, CreateDebate is a very different forum with potentially different userbase and linguistic styles (including syntactic and semantic differences with Reddit). Thus, in this section, we will report a real-world data-driven survey that establishes the validity of our detector on the CreateDebate dataset. 

\noindent\textbf{Study setup}: Our goal was to test the accuracy of our model output on the CreateDebate dataset. To that end, we ran our BERT-based detector on this dataset and randomly sampled $50$ comments which were classified as ad hominem by our detector and another $50$ comments which were non-ad hominem. Next, we created a simple online survey. We used Prolific\footnote{\url{https://www.prolific.co/}}, a crowdsourcing platform, to recruit participants for our survey.  We recruited $18+$ years old US nationals who were fluent in English, had a $100\%$ approval rating on the platform and had participated in at least $200$ previous studies. In this survey, we presented a set of comments (from our sample of $100$ random CreateDebate comments) along with a link to access the original CreateDebate discussion and its replies to each participant. Then we asked the participants to mark each of those comments as ad hominem or non ad hominem. For each comment, we additionally presented (in case the participant deem it to be an ad hominem) top three phrases identified by our model (with highest weights) and enquired if these key phrases indeed make this content ad hominem (the participants could also add their own phrases in a free form text field). This part of our survey was aimed to validate the explainability of our model. In total, $15$ participants gave three responses for each of the $100$ comments (each participant gave responses for a batch of $20$ comments); comparing the annotators across $5$ batches yielded substantial inter-annotator agreement ($0.73$ Fliess' $\kappa$). The average time per participant was $8$ minutes and we compensated them with \$$1$. The survey instrument is provided in Appendix~\ref{instruments}. 

\noindent\textbf{Results}: We found that for $94\%$ of CreateDebate comments, the labels given by participants were the same as the predicted label by our model, signifying the high validity of our model output even on the CreateDebate dataset. Furthermore, for $94.3\%$ of ad hominem comments, the key phrases identified by our model (with the highest attention scores) exactly matched with the participant-identified phrases. This shows the power of the generalizability of our model. 

\section{Characterizing ad hominem fallacy usage in CreateDebate discussions} \label{aggregate-analysis}

We used our (almost) accurate and explainable detector on the CreateDebate dataset and characterized the ad hominem fallacies. We will start by exploring the volume of ad hominem fallacies in the wild. 

\subsection{Usage of ad hominem fallacies in CreateDebate}

\if{0}
\begin{table}[!htbp]
\centering
\small
\begin{tabular}{lrrrr}
\hline
\textbf{Topic}      & \multicolumn{2}{c}{\textbf{Ad hominem comments}} & \multicolumn{2}{c}{\textbf{Ad hominem users}}      \\
                    & \textbf{(\#)} & \textbf{(\%)}            & \textbf{(\#)} & \textbf{(\%)}              \\ \hline
Politics            & 42,718        & 35.64                    & 2,686         & 34.95 \\
Religion            & 20,194        & 26.08                    & 1,712         & 37.52 \\
World News          & 6,701         & 24.44                    & 1,191         & 32.88 \\
Science             & 5,437         & 26.28                    & 996           & 35.11 \\
Law                 & 2,642         & 23.98                    & 482           & 33.57 \\
Technology          & 1,401         & 16.64                    & 676           & 25.28 \\
\hline
\textbf{Total}      & 79,093        & 29.97                    & 4,965         & 33.19 \\
\hline
\end{tabular}
\caption{Basic statistics of ad hominem content in our collected CreateDebate dataset. Users who have 50\% or more comments as ad hominem arguments are referred to as ad hominem users here. \new{The fifth column indicates the \% of such ad hominem users in the different topical forums.}}
\label{table:CreateDebateDatasetDescriptionAd}
\end{table}
\fi

\begin{table}[!htbp]
\centering
\small
\begin{tabular}{lrr}
\hline
\textbf{Topic}      &\textbf{\% AH comment} &\textbf{\% AH users}\\ \hline
Politics            & \new{$37.03 \pm 6.36$} & $34.95$ \\
Religion            & \new{$27.47 \pm 6.37$} & $37.52$ \\
World News          & \new{$25.21 \pm 5.30$}         & $32.88$ \\
Science             & \new{$27.90 \pm 6.10$}  & $35.11$ \\
Law                 & \new{$25.41 \pm 5.07$} & $33.57$ \\
Technology          & \new{$18.94 \pm 5.49$} & $25.28$ \\
\hline
\textbf{Overall}      & \new{$31.23 \pm 6.12$} & $33.19$ \\
\hline
\end{tabular}
\caption{Basic statistics of ad hominem (AH) content in our collected CreateDebate dataset. \new{Error bands for our estimates for ad hominem content are also included.} Users who have $50\%$ or more comments as ad hominem arguments are referred to as ad hominem users here. \new{The third column indicates the percentage of such ad hominem users in the different topical forums.}}
\label{table:CreateDebateDatasetDescriptionAd}
\end{table}

We simply run our BERT-based detector on CreateDebate data to find the answer to the question---do CreateDebate users leverage ad hominem fallacy? We present the answer in Table~\ref{table:CreateDebateDatasetDescriptionAd}. Surprisingly, the percentage of ad hominem comments in the CreateDebate forum is alarmingly high, especially for CreateDebate topical forums related to Politics \new{($37.03\%$)}. In fact, a large number of users are using these ad hominem fallacies---$34.95\%$ for Politics, demonstrating, ad hominem fallacies are used rampantly in the wild. These numbers contrast with the Reddit CMV forum where \citet{habernal-etal-2018-name} found only $0.02\%$ posts to be ad hominem. Even for a regular online discussion, only $19.5\%$ of comments under online news articles were found to be incivil \citep{Coe2014OnlineAU}, much lower than the reported fraction of ad hominems. We show some examples of topical ad hominems posted on CreateDebate in Table~\ref{table:AdHominemExamples}.

\begin{table}[!htbp]
\centering
\small
\begin{tabular}{cc|cc}
\hline
\textbf{Username} & \textbf{\# TLC}  & \textbf{Username} & \textbf{\# DR}  \\ \hline
\bf{UserA}             &$1,445$             &\bf{UserC}              & $5,371$\\
\bf{UserB}             &$1,388$             &\bf{UserB}              & $4,039$\\
\bf{UserC}             &$1,078$             &\bf{UserA}              & $2,694$\\
\bf{UserD}             &$1,077$             &UserK              & $2,513$\\
UserE             &$1,030$             &\bf{UserD}              & $2,365$\\
UserF             &$845$               &UserL              & $2,345$\\
\bf{UserG}             &$770$               &\bf{UserG}             & $1,577$\\
UserH             &$603$               &UserM              & $1,297$\\
\bf{UserI}             &$521$               &\bf{UserI}              & $1,200$\\
UserJ             &$506$               &UserN              & $1,069$\\
\hline
\end{tabular}
\caption{Users who posted most of the top-level comments (denoted by column \textincon{TLC}) and who received most of the direct replies (denoted by column \textincon{DR}) on the Politics sub-forum of CreateDebate. We omitted usernames for privacy concerns and common users from both the sides are represented in bold font.}
\label{table:Top10}
\end{table}

\begin{table}[!htbp]
\centering
\small
\begin{tabular}{lrrr}
\hline
\textbf{\# CC}   &\textbf{\% Users}  &\textbf{\% Comments}   &\textbf{\% Ad hominem} \\ \hline 
$\le 10$          &$87.8$               &$11.8$	                 &$16.8$ \\ 
$11-50$           &$8.1$                &$10.5$	                 &$24.6$ \\ 
$51-100$          &$3.6$                &$31.0$	                 &$26.2$ \\ 
$101-2000$        &$0.4$                &$20.2$	                 &$26.1$ \\ 
$\ge 2000$       &$0.1$                &$26.5$	                 &$55.2$ \\ 
\hline
\textbf{Total}            & $100.0$	         &$100.0$	                 &$37.03$ \\ 
\hline
\end{tabular}
\caption{Grouping users on the basis of their top-level comment count (denoted by column \textincon{\#CC}) and analyzing percentage of users and top-level comments for each group. Column \textincon{\%Ad hominem} denotes the percentage of ad hominem comments in the given group only.}
\label{table:GroupStatistics}
\end{table}


\begin{figure*}[t]
\centering
\begin{tabular}{ccc}
\includegraphics[width=0.33\textwidth]{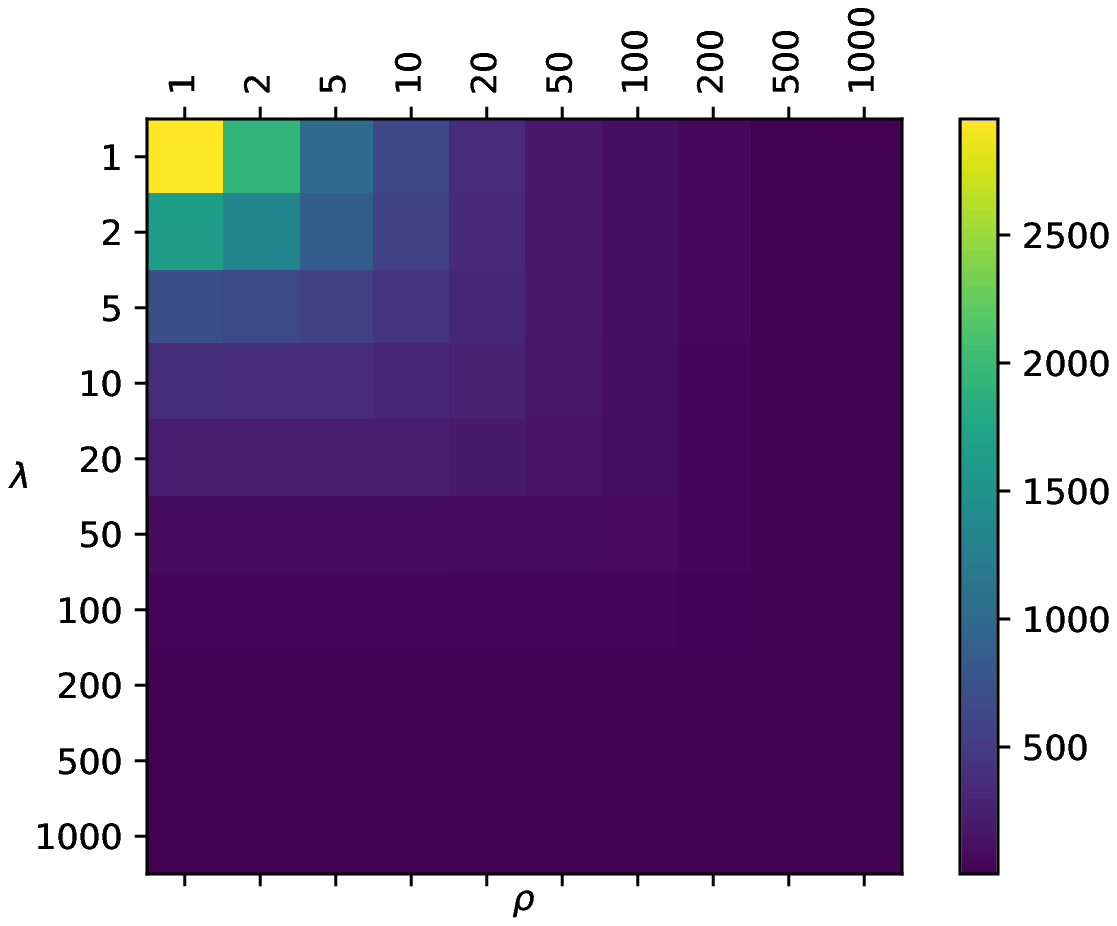} &
\includegraphics[width=0.33\textwidth]{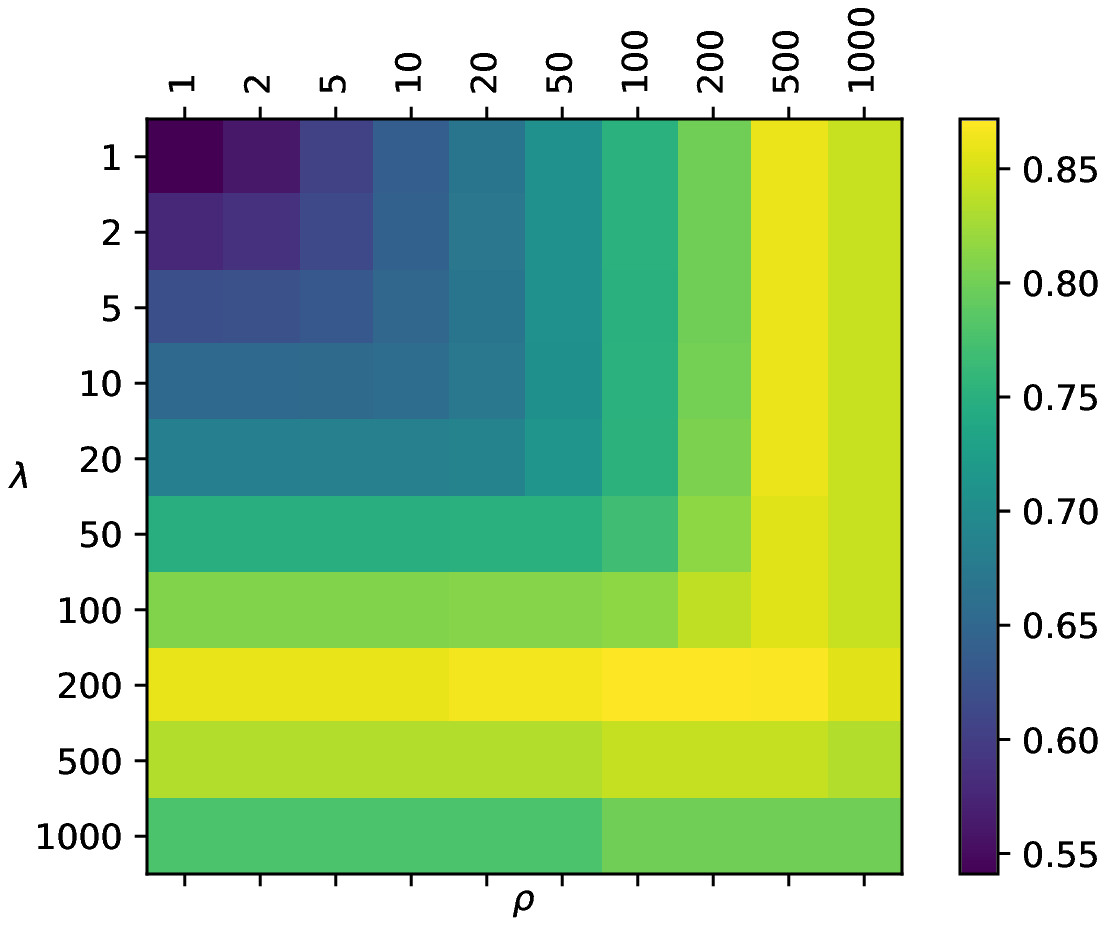} &
\includegraphics[width=0.33\textwidth]{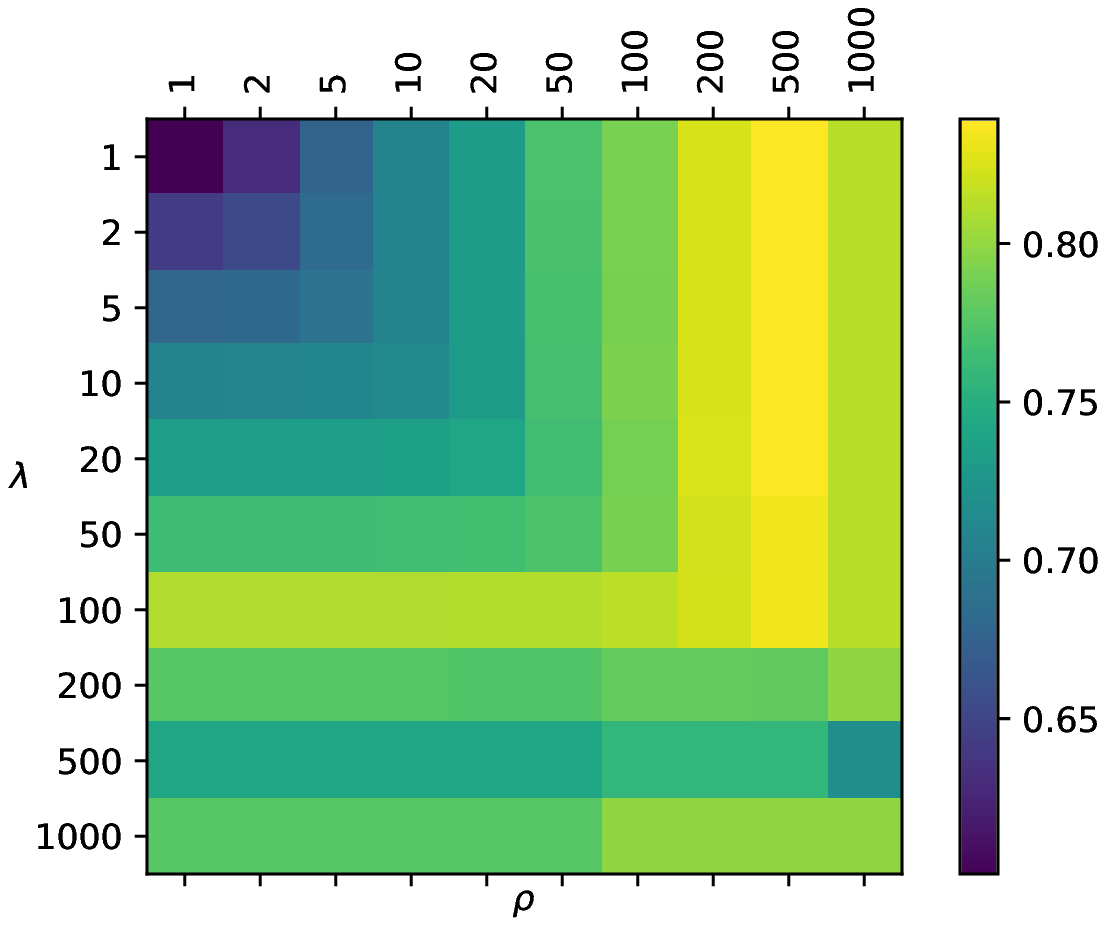} \\
(a)  & (b) & (c)  \\[6pt]
\end{tabular}
\caption{Variation in (a) number of authors in $S(\lambda,\rho)$; reciprocity amongst the users computed using (b) support network and (c) dispute network for different $\lambda$ and $\rho$ for CreateDebate Politics sub-forum.}
\label{fig:ReciprocityAggregate}
\end{figure*}


\begin{figure*}[htbp]
\centering
\includegraphics[width=\textwidth]{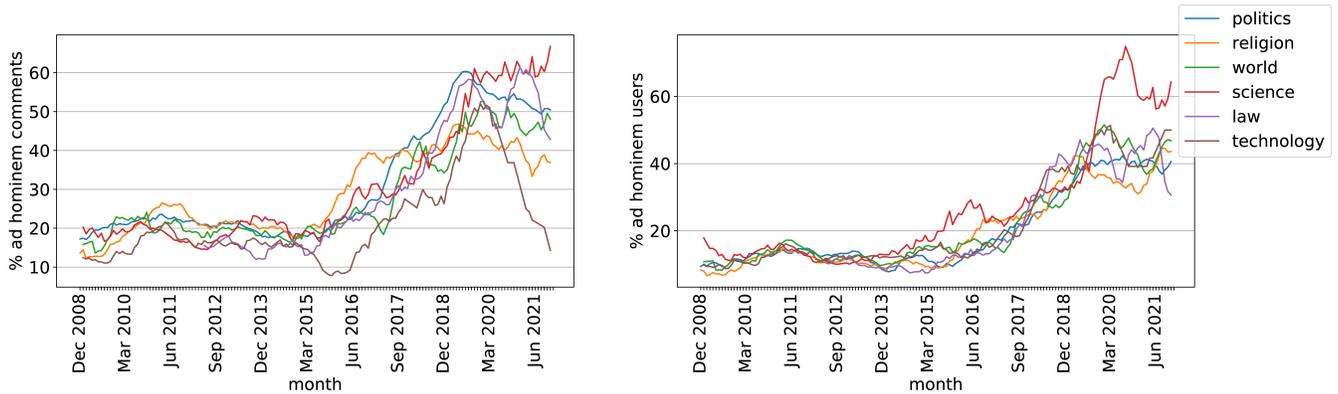}
\caption{Variation in percentage of ad hominem posts posted per month on CreateDebate across different topics \textit{(left)} and percentage of users for whom 50\% or more posts are ad hominem per month \textit{(right)} (averaged over 1 year).}
\label{fig:Temporal}
\end{figure*}


Now, we ask an obvious question---why is this percentage alarmingly high in contrast to CMV’s $0.02\%$, even though there is a mechanism to \textit{report} a comment on CreateDebate? 

To investigate, we focus on ad hominem posts from `Politics' subforum (owing to its $37.03\%$ ad hominem content). 

As the statistics in Table~\ref{table:Top10} show, the Politics sub-forum of CreateDebate follows a heavy tail distribution, where \new{the} $10$ most active users posted about $26\%$ of the top-level comments, it also shows the users who received the most direct replies to their posts on the subforum. Interestingly, both sides have $6$ usernames in common, signifying the possible influence of only a handful of key players. Thus, we ask---do these active users have any role in elevating the fraction of ad hominem arguments in the CreateDebate forum?




\subsection{Correlation between activity and ad hominem}

We wanted to investigate if a handful of users are colluding to upload disproportionately more ad hominem content. To that end, we define $X(\lambda)$ as the set of authors who posted at least $\lambda$ top-level comments, and $Y(\rho)$ as the set of authors who received at least $\rho$ direct replies to their comments. We define the set  $S(\lambda,\rho)$ as $X(\lambda) \cap Y(\rho)$ and attempt to understand the activity of users with different levels of $\lambda$ and $\rho$. 


\noindent\textbf{Highly active users act as a community while posting}: We start by creating directed graphs showing support and dispute between the authors in $S(\lambda,\rho)$. The weights of the edge from node $A$ to node $B$ in support and dispute networks represents how many times author $A$ agreed/disagreed with author $B$ via direct reply. As the results in Figure~\ref{fig:ReciprocityAggregate} show, the number of authors in the set decreases if either $\lambda$ or $\rho$ is increased, yet the reciprocity\footnote{\new{``Reciprocity'' is defined to be a measure of how likely a pair of vertices are mutually connected~\cite{reciprocity-defn} (i.e., both $node_1 \rightarrow node_2$ and $node_2 \rightarrow node_1$ edges exist) in a directed network. In our setup, this indicates, that if an author $a_1$ supports (disputes) author $a_2$ then what is the likelihood that $a_2$ would support (dispute) $a_1$.}} in support and dispute networks increases with an increase in $\lambda$ and $\rho$. Our finding implies that the influential actors, who also happened to write most posts and receive most replies (high $\lambda$ and $\rho$), participate in the debates not as an individual but as small-sized communities/cohorts of highly active users. 


\noindent\textbf{Community of highly active users post a disproportionately high volume of ad hominem posts}: We started with the hypothesis that the alarmingly high ad hominem content present in the CreateDebate forum is mostly generated by small communities of highly active users. These users always post together and reply to each other. To check whether our hypothesis is true, we grouped the users participating in political debates on the CreateDebate forum based on their total comment count. Next, we test the fraction of ad hominem comments in the content generated by each group. The result is shown in Table~\ref{table:GroupStatistics}. The set of authors who wrote more than $2000$ comments constituted only $0.1\%$ of the users, yet they post around $26.5\%$ of total content with as high as $55.2\%$ of their content flagged as ad hominem. This group's activity is in stark contrast with the users who post less than $10$ comments---they posted only $11.8\%$ comments and a meager $16.8\%$ of those comments were ad hominem. This finding confirms our hypothesis and identifies an intriguing pattern of ad hominem posting---these illogical personal attacks are often used by a highly active community or cohort of users, presumably to shut down voices of less active opponents.

\begin{figure*}[htb]
\centering
\begin{tabular}{ccc}
\includegraphics[width=0.3\textwidth]{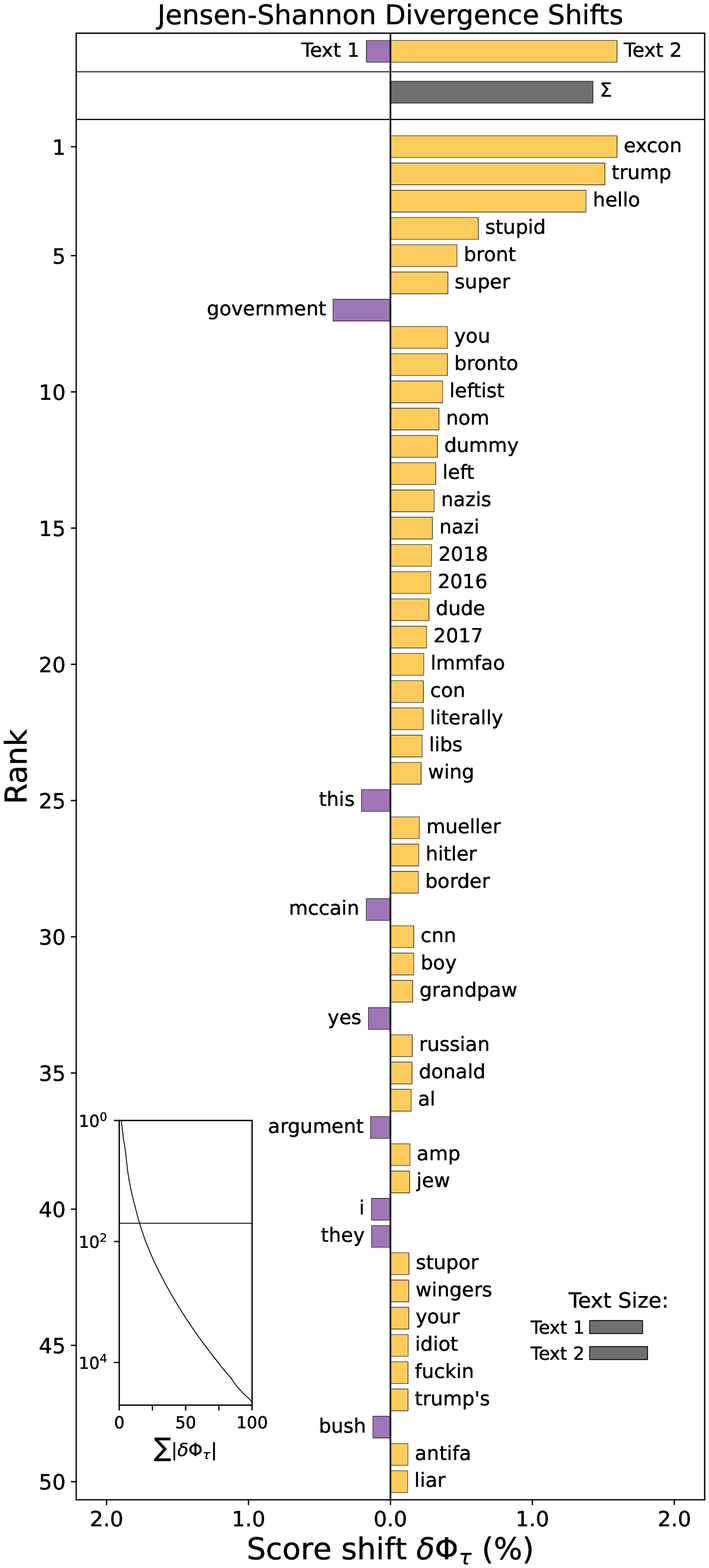} &
\includegraphics[width=0.3\textwidth]{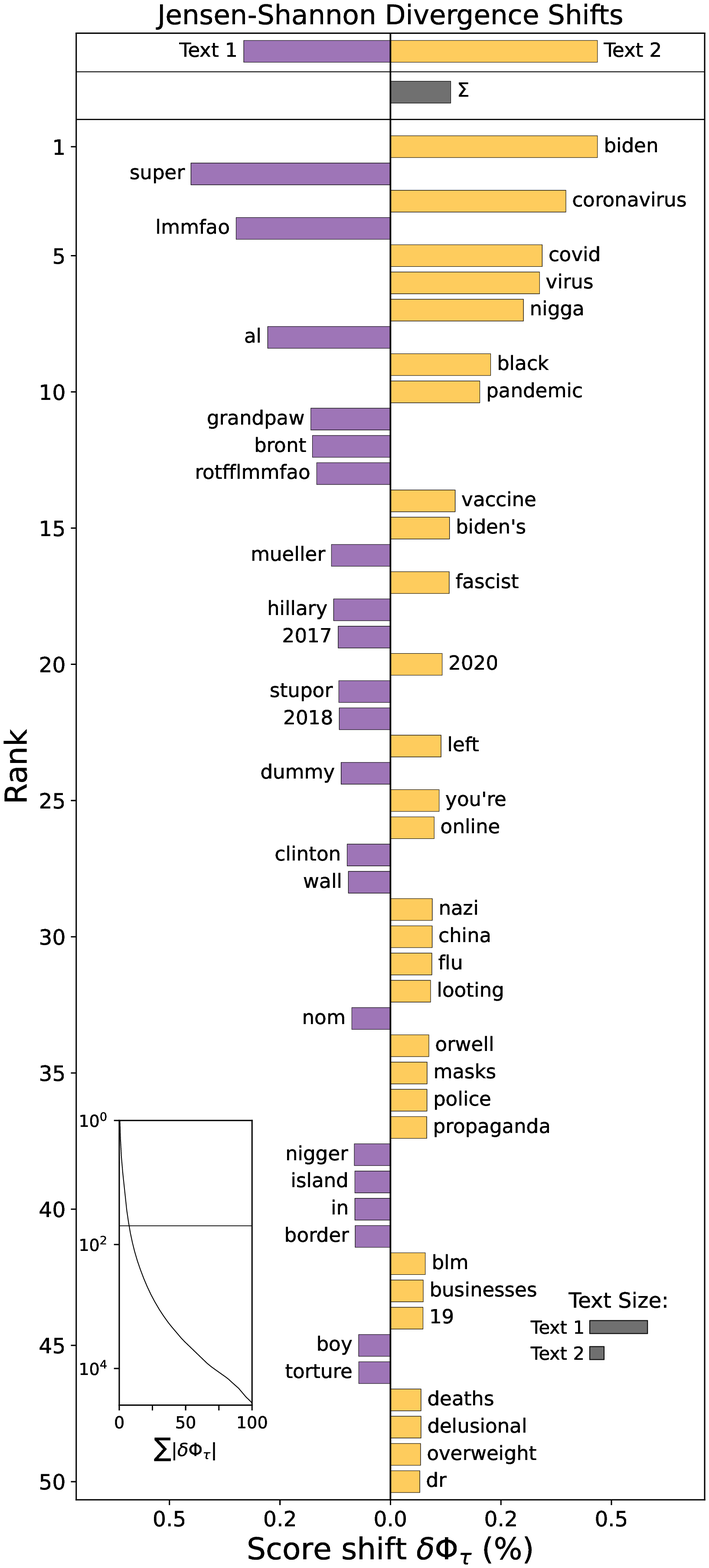} &
\includegraphics[width=0.3\textwidth]{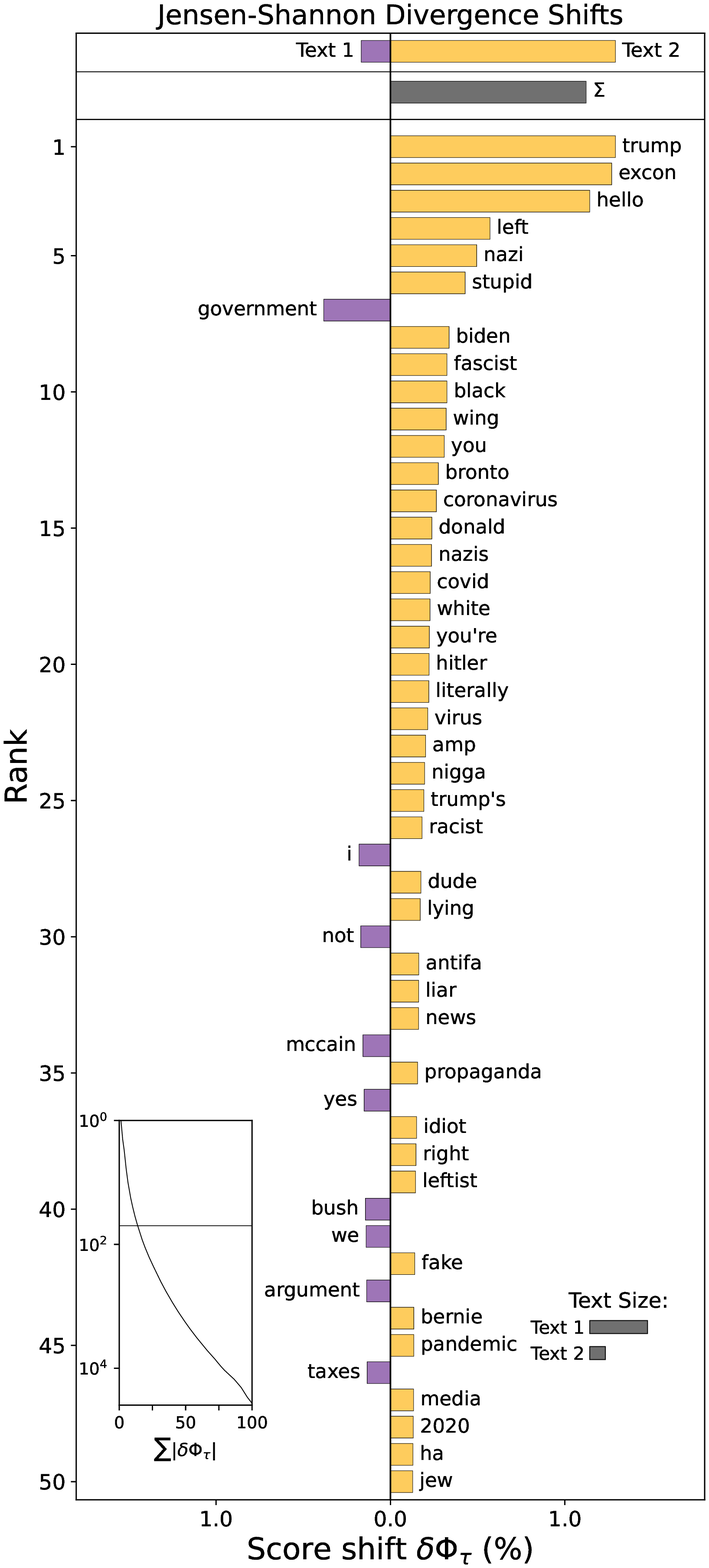} \\
(a)  & (b) & (c)  \\[6pt]
\end{tabular}
\caption{Word shift graphs constructed using Jensen-Shannon divergence for comparing sub-corpora (a) $\mathcal{H}_1$ and $\mathcal{H}_2$, (b) $\mathcal{H}_2$ and $\mathcal{H}_3$, and (c) $\mathcal{H}_1$ and $\mathcal{H}_3$. Ad hominem triggering words are prominent in $\mathcal{H}_2$ when compared to $\mathcal{H}_1$. For $\mathcal{H}_3$, these triggers are not prominent.}
\label{fig:WordShiftGraphs}
\end{figure*}

\noindent\textbf{Summary:} CreateDebate dataset contains a surprisingly high volume of ad hominem---$31.23\%$, which is much higher than the figures reported in earlier works on ad hominem content in other forums. Using the Politics subforum, we further showed that these ad hominem comments are largely facilitated by highly active CreateDebate users who collude among themselves to post together on the same discussion-reply threads and reply to each other\footnote{Our observations also hold for other topics; these results are not shown for brevity.}. 

\subsection{\new{Characterising users posting ad hominem comments}}

\new{A natural question at this point is whether it is possible to characterise, early in time, the users who have a tendency to post ad hominem comments. In this subsection, we show that characteristic differences exist between those who post ad hominem comments and those who do not. To this end, we utilise the profile page of the users maintained by CreateDebate. The profile page contains various information about an user. These include the \textit{reward points} earned by a user, their \textit{efficiency} while debating, the \textit{number of debates} they participated in, the \textit{number of comments} they posted since they joined the forum. Other additional features consist of \textit{allies, enemies} or \textit{hostiles} 
of any given user\footnote{\new{Allies are users with whom one has common interests and opinions; enemies are the ones with whom a user has different interests and opinions; a hostile is one who declared the user as their enemy. Please check CreateDebate FAQ page for more information.}}.}

\begin{table*}[]
\centering
\small
\begin{tabular}{p{0.25\linewidth} p{0.20\linewidth} p{0.20\linewidth} p{0.15\linewidth}}
\hline
\textbf{Characteristics}   &\textbf{Class $\mathcal{C}_1$}  &\textbf{Class $\mathcal{C}_2$} & \textbf{MWU $p$ value}    \\ \hline 
\textbf{\#} posts         &$49.26 \pm 294.71$             &$2.02 \pm 2.25$	                 &\num{0}\\ 
reward points             &$169.41 \pm 1070.35$           &$6.89 \pm 14.31$	                 &\num{0} \\ 
efficiency                &$88.31 \pm 13.07$              &$91.58 \pm 13.55$	             &\num{1.86e-142} \\ 
\textbf{\#} allies        &$1.84 \pm 9.11$                &$0.14 \pm 0.83$	                 &\num{2.76e-249} \\ 
\textbf{\#} enemies       &$0.66 \pm 6.21$                &$0.03 \pm 0.24$	                 &\num{3.67e-203} \\ 
\textbf{\#} hostiles      &$0.59 \pm 2.99$                &$0.03 \pm 0.20$	                 &\num{0} \\ 
reciprocity (SN)          &$0.64 \pm 0.06$                &$0.58 \pm 0.02$	                 &\num{3.18e-264} \\ 
reciprocity (DN)          &$0.69 \pm 0.05$                &$0.64 \pm 0.03$	                 &\num{1.22e-263} \\ 
\hline
\end{tabular}
\caption{\new{Average and standard deviation of different characteristics along with $p$ value computed using Mann–Whitney U test for the distribution of the two classes $\mathcal{C}_1$ (users who have posted at least one ad hominem comment) and $\mathcal{C}_2$ (users who have not posted any ad hominem comment). Low $p$ value for MWU test denotes that the distributions of the two classes are statistically very different for each other across all user characteristics. \textit{SN} denotes support network and \textit{DN} denotes dispute network.}}
\label{table:UserChar}
\end{table*}

\if{0}
\begin{table*}[]
\centering
\small
\color{blue}
\begin{tabular}{p{0.25\linewidth} p{0.20\linewidth} p{0.20\linewidth} p{0.15\linewidth}}
\hline
\textbf{Characteristics}   &\textbf{Class $\mathcal{C}_1$ ( $>$ 50\%)}  &\textbf{Class $\mathcal{C}_2$ (= 0\%)} & \textbf{MWU $p$ value}    \\ \hline 
\textbf{\#} posts         &$26.77 \pm 314.2$             &$2.02 \pm 2.25$	                 &\num{6.50e-8}\\ 
reward points             &$58.33 \pm 645.24$           &$6.89 \pm 14.31$	                 &\num{1.10e-3} \\ 
efficiency                &$88.31 \pm 15.52$              &$91.58 \pm 13.55$	             &\num{6.65e-21} \\ 
\textbf{\#} allies        &$0.28 \pm 2.03$                &$0.14 \pm 0.83$	                 &\num{6.00e-3} \\ 
\textbf{\#} enemies       &$0.42 \pm 7.78$                &$0.03 \pm 0.24$	                 &\num{1.28e-34} \\ 
\textbf{\#} hostiles      &$0.39 \pm 3.12$                &$0.03 \pm 0.20$	                 &\num{7.68e-147} \\ 
reciprocity (SN)          &$0.62 \pm 0.06$                &$0.58 \pm 0.02$	                 &\num{9.14e-37} \\ 
reciprocity (DN)          &$0.68 \pm 0.06$                &$0.64 \pm 0.03$	                 &\num{3.12e-37} \\ 
\hline
\end{tabular}
\caption{\new{Average and standard deviation of different characteristics along with $p$ value computed using Mann–Whitney U test of the distribution of the two classes $\mathcal{C}_1$ and $\mathcal{C}_2$. \textit{SN} denotes support network and \textit{DN} denotes dispute network.}}
\label{table:UserChar1}
\end{table*}
\fi

\if{0}
\begin{table*}[]
\centering
\small
\color{blue}
\begin{tabular}{p{0.25\linewidth} p{0.20\linewidth} p{0.20\linewidth} p{0.15\linewidth}}
\hline
\textbf{Characteristics}   &\textbf{Class $\mathcal{C}_1$ ( $>$ 50\%)} &\textbf{Class $\mathcal{C}_2$ ( $\le$ 50\% \& $>$ 0\%)}  &\textbf{Class $\mathcal{C}_3$ (= 0\%)}     \\ \hline 
\textbf{\#} posts         &$26.77 \pm 314.2$              &$60.52 \pm 283.79$	                 &$2.02 \pm 2.25$\\ 
reward points             &$58.33 \pm 645.24$             &$224.95 \pm 1225.17$	                 &$6.89 \pm 14.31$	 \\ 
efficiency                &$88.31 \pm 15.52$              &$88.31 \pm 11.65$	             &$91.58 \pm 13.55$ \\ 
\textbf{\#} allies        &$0.28 \pm 2.03$                &$2.62 \pm 10.98$	                 &$0.14 \pm 0.83$ \\ 
\textbf{\#} enemies       &$0.42 \pm 7.78$                &$0.78 \pm 5.25$	                 &$0.03 \pm 0.24$ \\ 
\textbf{\#} hostiles      &$0.39 \pm 3.12$                &$0.69 \pm 2.93$	                 &$0.03 \pm 0.20$ \\ 
reciprocity (SN)          &$0.62 \pm 0.06$                &$0.64 \pm 0.06$	                 &$0.58 \pm 0.02$ \\ 
reciprocity (DN)          &$0.68 \pm 0.06$                &$0.70 \pm 0.05$	                 &$0.64 \pm 0.03$ \\ 
\hline
\end{tabular}
\caption{\new{Average and standard deviation of different characteristics along with $p$ value computed using Mann–Whitney U test of the distribution for the two classes $\mathcal{C}_1$ (users who have posted at least one ad hominem comment) and $\mathcal{C}_2$ (users who have not posted any ad hominem comment). Low  \textit{SN} denotes support network and \textit{DN} denotes dispute network.}}
\label{table:UserChar2}
\end{table*}
\fi

\new{We collected these characteristics for all the users of CreateDebate and partitioned the users into two classes---those who have posted ad hominem comment at least once on the forum ($\mathcal{C}_1$) and those who haven't ($\mathcal{C}_2$), and then computed the average and standard deviation of these characteristics for these two classes. We also considered the average and standard deviation of reciprocity observed from the support and dispute networks.

\review{We observed that users belonging to class $\mathcal{C}_1$, on average, tend to post more often, have more reward points and less efficiency while debating, have more number of allies, enemies and hostiles and also show higher reciprocity, both in support and dispute network, when compared to the users belonging to class $\mathcal{C}_2$.} The results are shown in Table~\ref{table:UserChar}. It can be observed that the distributions of the two classes are statistically very different from each other (very low $p$-values using Mann-Whitney U test). In a future work, one can featurize these user profile based characteristics to build an ML model to predict the propensity of a user to post ad hominem comments early in time. However, development of such models would, in turn, need a detailed understanding of the temporal evolution of ad hominem usage in the platform. So, in the next section, we investigate the following---does the huge prevalence of ad hominem has any correlation with time, or was ad hominem always equally prevalent in the platform? } 

\section{Understanding temporal variations in ad hominem usage} \label{temporal-analysis}

\begin{figure*}[htb]
\centering
\begin{tabular}{cccc}
\includegraphics[width=0.24\textwidth]{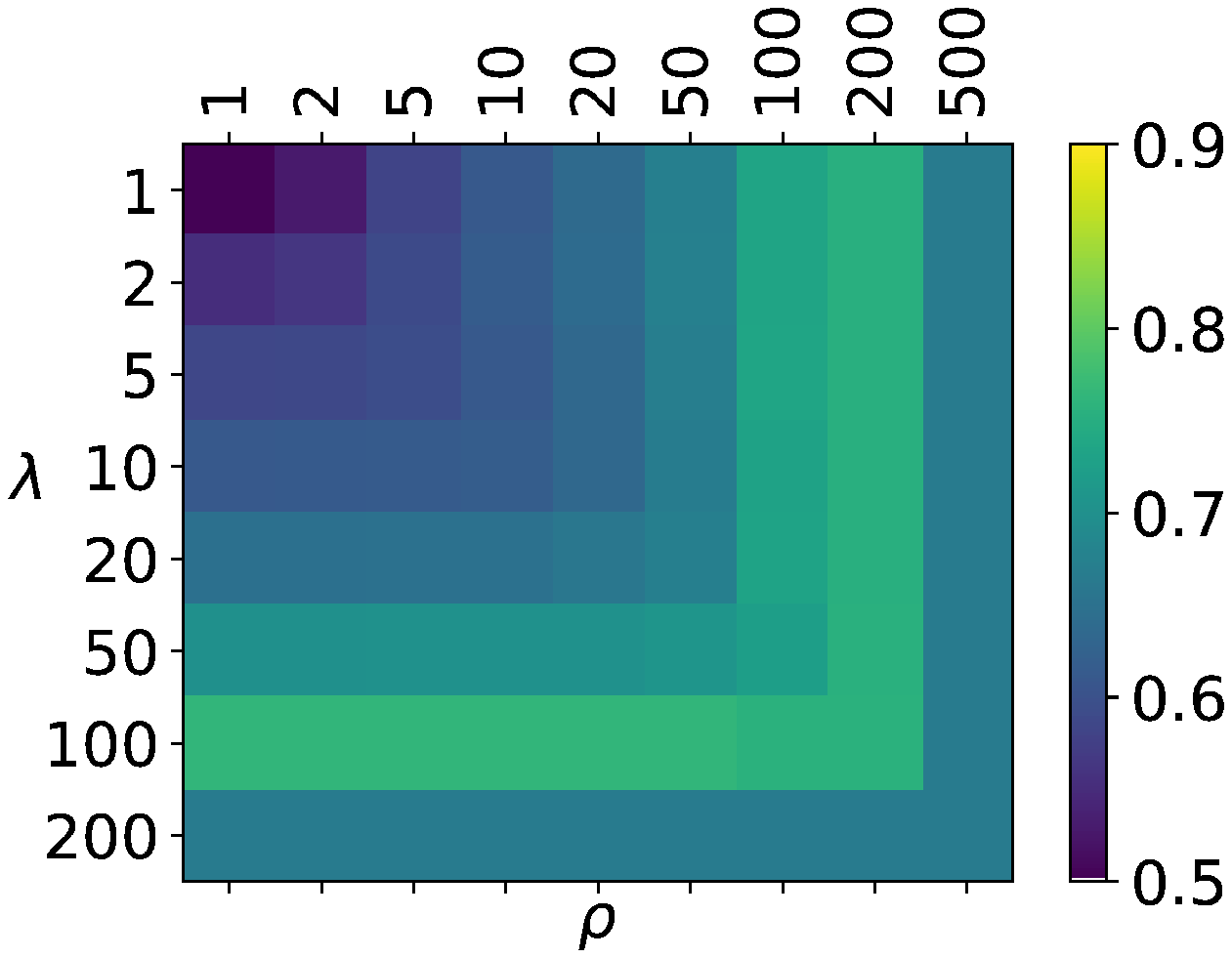} &
\includegraphics[width=0.24\textwidth]{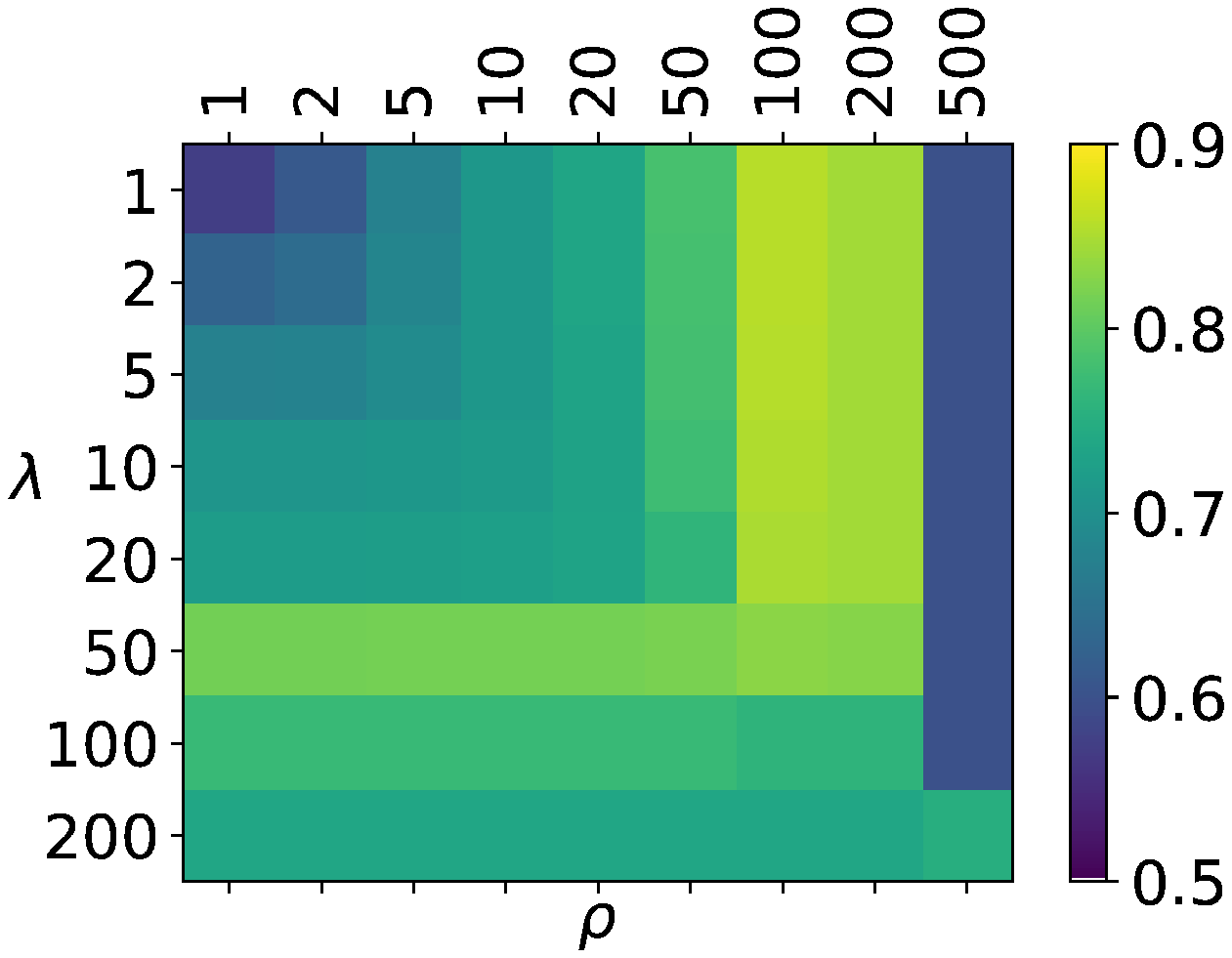} &
\includegraphics[width=0.24\textwidth]{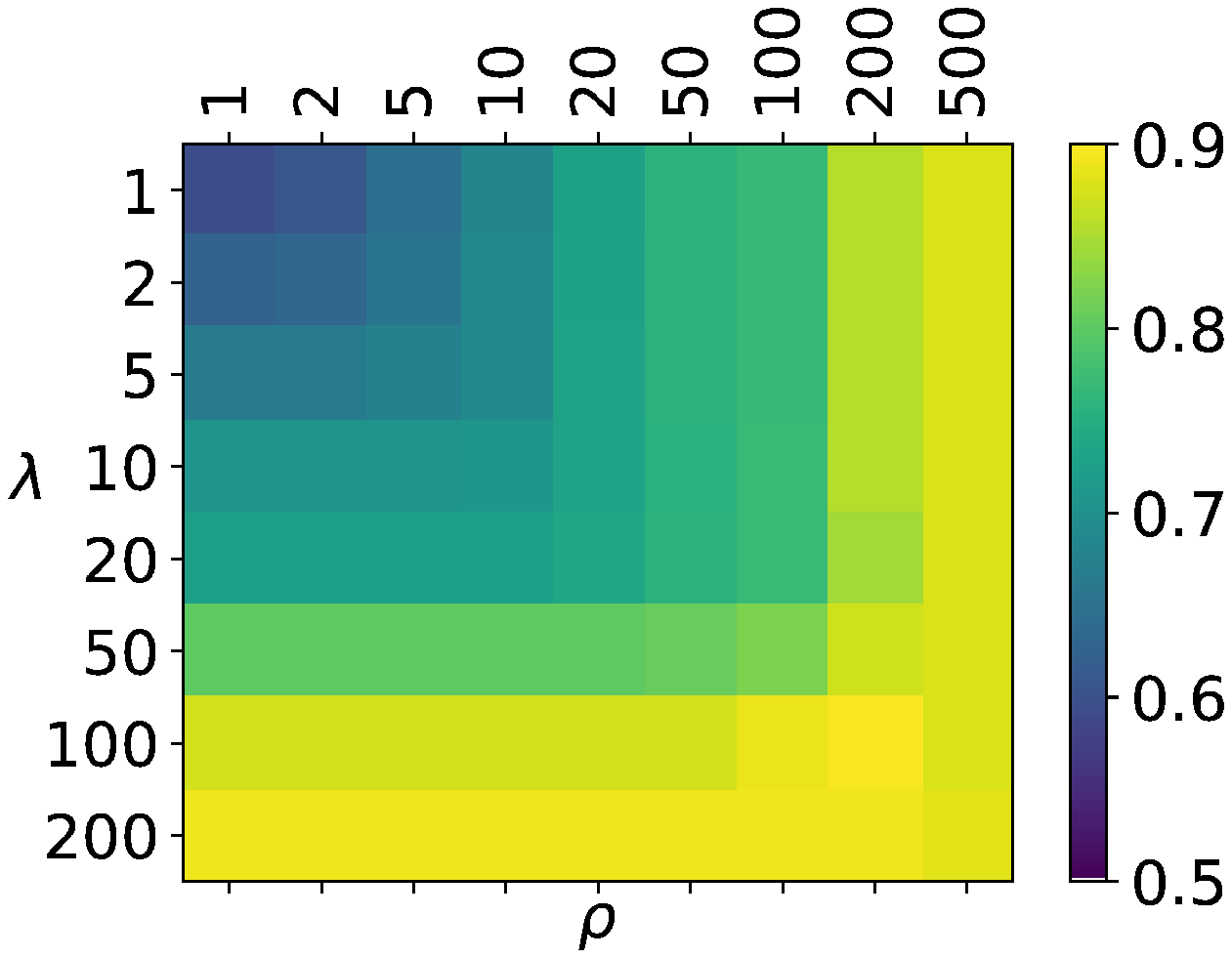} &
\includegraphics[width=0.24\textwidth]{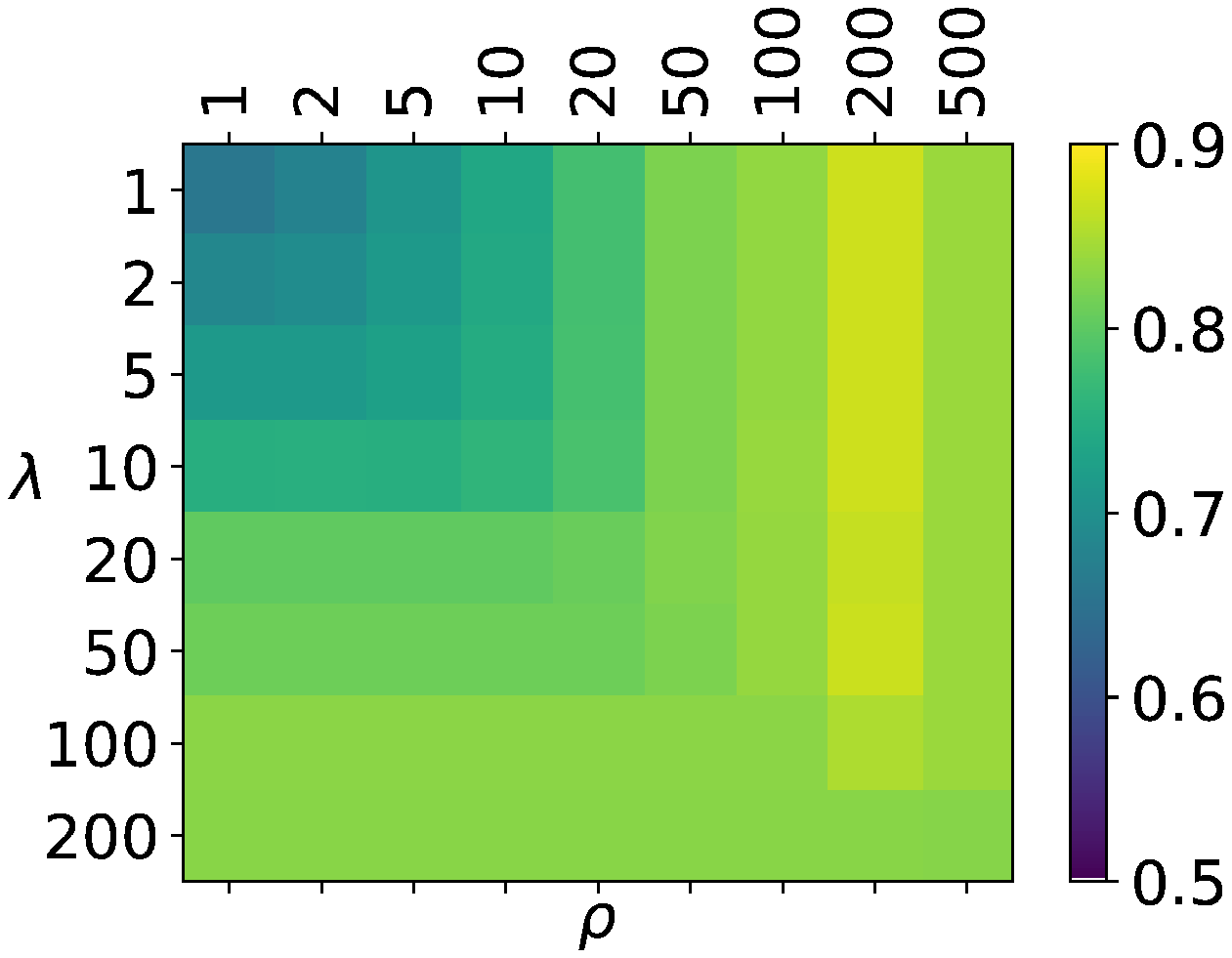} \\
(a) & (b) & (c) & (d) \\[6pt]
\end{tabular}
\caption{Reciprocity amongst the users of Politics sub-forum of CreateDebate computed using (a) support network and (b) dispute network constructed from sub-corpora $\mathcal{H}_1$, and (c) support network and (d) dispute network constructed from sub-corpora $\mathcal{H}_2$. $\lambda$ and $\rho$ are also varied. Reciprocity has increased significantly between the users for sub-corpora $\mathcal{H}_2$ when compared to $\mathcal{H}_1$.}
\label{fig:ReciprocityTemporal}
\end{figure*}

CreateDebate was launched in $2008$ as a tool ‘that democratizes the decision-making process through online debate’. However, as we have observed, the percentage of ad hominem content
\new{on the website}
is alarmingly high
\new{when}
compared to any
\new{other}
regular debate forum. So, what went wrong? To answer this question, we perform the first temporal analysis of \new{ad hominem usage}
for the CreateDebate forum. Our first task involved generating temporal snapshots to capture the month-wise activity on the site. As our dataset spans from February $2008$ to November $2021$, it would be very clumsy to do the day-wise activity analysis, and a year-wise scheme will have only $14$
\new{data}
points. Hence, we chose to study month-wise activities.
\new{The variation in the percentage of comments which were flagged as ad hominem and the percentage of users who had been posting such comments is shown in Figure~\ref{fig:Temporal} for each month in the period.}
\new{It can be observed that the plots for all the topical sub-forums follow a similar trend---initially they are stationary, then they show a steep rise and then they fall.}
\new{In order to gain insights of what exactly triggered this sharp rise and fall, we performed change point detection experiments \citep{TRUONG2020107299} to quantitatively partition the corpus into three sub-corpora---the stationary $\mathcal{H}_1$, the rise $\mathcal{H}_2$ and the fall $\mathcal{H}_3$. We used variation in the number of comments posted, percentage of comments which were flagged as ad hominem and percentage of users who were posting such comments for each month across all topical sub-forums as input to the change point detection algorithm which uses dynamic programming to find the optimal partition using RBF kernels as cost function. The cutoffs for the partitions as predicted by the algorithm are \textit{March $2017$} and \textit{September $2019$}. Hence, the timeline for the sub-corpora are---$\mathcal{H}_1$ (February $2008$ -- February $2017$), $\mathcal{H}_2$ (March $2017$ -- September $2019$) and $\mathcal{H}_3$ (October $2019$ -- November $2021$).}

We
\new{then}
compared these
\new{sub-}corpora
by generating word-shift graphs using Jensen-Shannon divergence,
\new{which are shown}
in Figure~\ref{fig:WordShiftGraphs}. It can be observed that the use of terms that act as triggers to ad hominem argumentation is very prominent in
\new{$\mathcal{H}_2$ when}
compared to
\new{$\mathcal{H}_1$}.
These triggers are also present in
\new{$\mathcal{H}_3$},
but they are not as dominant as in
\new{$\mathcal{H}_2$}.
We constructed the support and the dispute networks
\new{for $\mathcal{H}_1$ and $\mathcal{H}_2$} (see Figure~\ref{fig:ReciprocityTemporal}) and observed that the reciprocity has increased significantly between users who wrote at least
\new{$100$}
top-level comments or received at least
\new{$200$}
direct replies, for support as well as dispute networks.

\new{It is very interesting to note that the timelines for the $2016$ US Presidential election and the Covid-$19$ outbreak are very close to the predicted cutoffs of the partitions.}
We observed that the CreateDebate forum was heavily used for political debates during the $2016$ US Presidential election. Our hypothesis is that usage of ad hominem argumentation was accelerated after the $2016$ US Presidential debates – forum becoming highly polar, people choosing sides. As it has been observed throughout history, the use of illogical arguments is very prevalent when people are discussing Politics, but to win debates, the use of
\new{ad hominem comments}
skyrocketed
\new{on the forum}.
However,
\new{due to}
the Covid-$19$
pandemic, the activity of users on the forum declined significantly,
\new{thus curbing the ad hominem usage to some extent.}

\begin{figure}[htb]
\centering
\includegraphics[width=\linewidth]{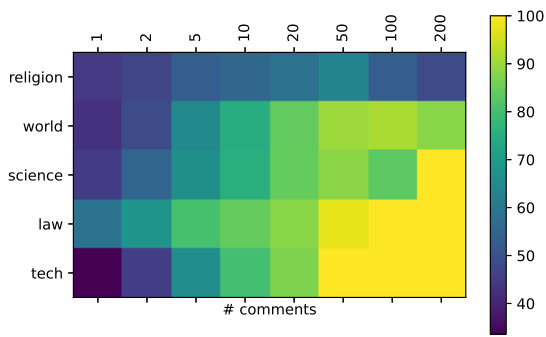}

\caption{Overlap (\%) of user base for different topical sub-forums with Politics sub-forum on CreateDebate. The overlap is computed across users with different comment count (denoted here as \textincon{\#comments}). The heat map shows that the overlap is very large across different sub-forums, and it approaches 100\% if we only consider the highly-active users.}
\label{fig:OverlapUser}
\end{figure}

This is a plausible explanation about rising ad hominem arguments in political debates. But, what about other topical forums like Religion? Why is the ad hominem content increasing for these topics? To understand this complex phenomenon, we partitioned the religious debates using the above scheme and peeked into what users are talking about. We observed that the religious debates that were published before the 
$2016$ US Presidential debates have a negligible political angle, however, those published after the Presidential debates were highly convoluted with Politics. As the results in Figure~\ref{fig:OverlapUser} show, each topical
\new{sub-forum}
on the CreateDebate site has a huge user overlap with Politics, especially the highly active users. For categories like Science and Law, the overlap approaches $100\%$. This explains why the increase in ad hominem comments and users posting them across different categories show similar trends as Politics (see Figure~\ref{fig:Temporal}). So, it is likely that political discussions are root cause of the alarmingly high ad hominem content on CreateDebate, which skyrocketed after start of the $2016$ US Presidential debates.

\section{Concluding discussion}\label{sec:implications}

\noindent In this work, for the first time, we shed a data-driven light on ad hominem usage in the wild by leveraging the CreateDebate data posted by tens of thousands of users over a period of more than a decade. We reported creating detectors with high accuracy whose judgment matched that of users for $94\%$ of the cases. Using this detector, we uncover that very surprisingly around \textit{one-third} of all content on CreateDebate is simply ad hominem. We deep-dived to find that a cohort of highly active users is responsible for this high fraction of ad hominems. Moreover, users particularly influenced by the political discourse resorted to this fallacy. While almost all data-driven studies suffer from intrinsic bias, we strongly believe this work is still valuable for understanding the ecosystem of cyber aggression as well as designing novel and more respectful debating platforms. In this final section, we will discuss the limitations as well as key implications of our work.

\subsection{Limitations}

Our work has a couple of limitations. First, our detector is bound by the annotation quality and volume of the CMV dataset. We believe we could have achieved higher accuracy with more data. However, even our detector achieved a significantly high accuracy compared to the prior works and we established that this detector is also valid on CreateDebate---a different forum and dataset, underpinning the efficacy of the detector. Second, our results might or might not be generalizable beyond Reddit, CreateDebate and in general online forums to share opinions and debate. Even then, it has a lot of societal importance since, as prior works show, such debate forums in themselves are extremely important to shape the Internet and public opinion~\citep{doi:10.1177/20563051211019004}. Third, our findings are often \new{correlations} and not \new{causations}. However, we believe these correlations show underlying large-scale behavioral patterns which promote ad hominem comments and, in effect, a toxic culture. Thus, in spite of this limitation, our work is still useful as a first attempt to shed light on ad hominem usage patterns.

\subsection{Implications}

We identify three key implications of our work for platform designers as well as platform regulators. 

\noindent \textbf{Defending against logical fallacies is becoming important}: One very surprising and concerning finding is that our dataset from CreateDebate, a popular opinion-forming and debating forming forum, is filled with ad hominem fallacies. This finding highlights the importance of understanding and defending against logical fallacies which perpetuate toxic culture and attempts to stop any opposing arguments using irrelevant personal attacks. Thus, to counter cyber-aggression and bring back respect to online spaces, the platforms should acknowledge this issue and actively design defenses against such fallacies. The key focus today is on defending against hate speech and misinformation. \new{The alarming rate of growth of ad hominems necessitates the design and development of dedicated countermeasures for this form of online malice.} 

\noindent \textbf{It is possible to detect and defend against these fallacies via automated means}: Our work demonstrated that we can leverage current extremely powerful techniques like BERT and GAN-BERT to create highly accurate ad hominem detectors, even when only a handful of annotated posts are available. Thus, our work can also be interpreted as a very strong proof of concept about using automated means (e.g., classifiers) by the platforms to protect against such fallacies along with hate speech and misinformation. 

\noindent \textbf{Users need to be nudged to reduce the usage of these fallacies}: Finally, the results of this work strongly hint at the need of nudging users to reduce logical fallacies. As our results show, highly active user cohorts in CreateDebate are using ad hominem in more than $50\%$ of their comments. \new{Hence, regular online awareness campaigns could be organised to urge users to report about any such suspicious behavior that come to their notice.} Furthermore, the substantially high ad hominem usage was rooted in the political climate of 2016 and now, it is spreading through other topics and forums, polluting the online space. The high volume of the affected population possibly even \new{hints} that many of these users might not even realize that they are utilizing fallacious arguments. Thus, the current platforms should focus on nudging the users against the usage of potential ad hominems even before they upload a fallacious post. \new{Using the models discussed in this paper, online debate forums and social media sites can nudge users when they are about to post ad hominem comments by making them aware of the ad hominem triggers present in their post. Moderators can also decrease exposure of such comments by pushing them at the bottom of the thread, and even flagging them. } Overall, we strongly believe that our findings will help policymakers and platform developers help detect and defend against ad hominem fallacies in online opinion influencing forums.

\bibliography{main}

\appendix


\section{Survey instrument} \label{instruments}

This section contains the survey instrument that we used during annotation studies.

\subsection{Instructions}
\tiny {

\noindent Identifying Personal Attacks in Comment Chains



\noindent An ad hominem argument (or argumentum ad hominem in Latin) is used to counter another argument. However, it's based on feelings of prejudice (often irrelevant to the argument), rather than facts, reason, and logic. An ad hominem argument is often a personal attack on someone's character or motive rather than an attempt to address the reasoning that they presented. 

\noindent Sometimes, people utilize ad-hominem argument (fallacy) because they want to appeal to other's emotions rather than their reasoning (since they are based on personal attack). Ad-hominem is often used in toxic conversations or comment chains in the internet.

\subsection{Examples}

\noindent Let's review several ad hominem examples. Unfortunately, they're prevalent in the courtroom and in politics, so we'll begin there. To no surprise, ad hominem arguments also occur in any sort of daily interaction, so we'll review a few more everyday examples, too. 

\noindent The more you read about examples of ad hominem arguments, the more you'll be able to spot them and, if need be, defend yourself against such arguments.

\noindent \textcolor{gray}{Next, gave five examples of Ad hominems identified from prior work in four situations---In the Court, In the Political Debates, Used in the Media, In Everyday Conversations}. 

\subsection{Task}


\noindent In this task, you will be shown 20 comments, one comment per page. For each comment, you will be asked whether the given comment is ad hominem argument or not. For additional context, each comment is provided with an URL of the full conversation (post and comments). You will also be asked to select some keywords from the comments shown, which you think, best describes your judgment (ad hominem or otherwise).

\noindent \textbf{Note} -- Devices you can use to take this study: Desktop and Tablet

\noindent \textcolor{gray}{For each of the 20 comments show the following}

\begin{itemize}
    \item Show the comment excerpt (with a link to the conversation for added context)
    
    \item Do you think this is an ad-hominem comment? $\bigcirc$ Yes $\bigcirc$ No
    
    \item \noindent \textcolor{gray}{If participant chose ad hominem} Select the phrases from the comments, which you think, makes it an ad hominem comment. If some other phrase makes it ad hominem, please enter that in 'Other' option. 
    $\square$ word 1 $\square$ word 2 $\square$ word 3  $\square$ other \_\_\_\_
\end{itemize}

}

\end{document}